\documentclass[10pt,twocolumn,letterpaper]{article}

\usepackage[pagenumbers]{cvpr} 

\definecolor{cvprblue}{rgb}{0.21,0.49,0.74}
\usepackage[pagebackref,breaklinks,colorlinks,allcolors=cvprblue]{hyperref}

\def\paperID{20225} 
\def\confName{CVPR}
\def\confYear{2026}

\usepackage{bm}
\usepackage{amssymb}
\usepackage{booktabs} 
\usepackage{afterpage} 
\usepackage{amsmath}
\usepackage[table]{xcolor} 
\usepackage{makecell}
\usepackage{multirow} 
\usepackage{xcolor}
\newcommand{\graytext}[1]{{\fontsize{6pt}{6pt}\selectfont\textcolor{gray}{#1}}}

\title{Bootstrap Your Own AV-Proxies: Adaptive Contrastive and Prototype Learning for Audio-Visual Segmentation}

\author{Junbo Zhang$^{1}$
\quad
Hang Su$^{1}$
\quad
Zhaofan Li$^{1}$
\quad
Hang Dong$^{2}$
\quad
Chao Sun$^{1 \dagger}$\\
$^{1}$School of Computer Science, Wuhan University\\
$^{2}$Academy of Advanced Interdisciplinary Studies, Wuhan University\\
{\tt\small \{junboz, chaosun\}@whu.edu.cn}
}

\begin{document}
\maketitle

\begingroup
\renewcommand\thefootnote{}
\footnotetext{† Corresponding author}
\endgroup

\begin{abstract}
Audio-Visual Segmentation (AVS) aims to accurately segment sounding objects in video frames by leveraging audio-visual correspondence cues. However, it remains challenging due to the intrinsic semantic incompleteness within a single modality and the semantic gap between audio and visual representations. Traditional feature-fusion decoding approaches struggle to suppress fusion noise effectively, while recent methods that incorporate data-dependent priors often increase the complexity of modeling audio-visual correlations, leading to poor cross-domain generalization. To address these issues, we propose a novel adaptive contrastive and prototype learning framework, BYOAVP, for AVS. Specifically, we design a Self-Supervised Audio Enhancement (SSAE) module that introduces contrastive learning to adaptively align audio representations with gradient-blocked visual embeddings, thus narrowing the semantic gap between modalities. Furthermore, a Dynamic Prototype Constraint (DPC) module is developed to refine pixel-wise category perception via momentum-based prototype updating, while enhancing localization of sounding regions through cross-modal interaction. Extensive experiments demonstrate that our method achieves SOTA performance across two AVS benchmarks and six sub-tasks, exhibiting strong robustness and generalization ability.

\end{abstract}    
\begin{figure}[ht]
   \centering
   \includegraphics[width=0.85\linewidth]{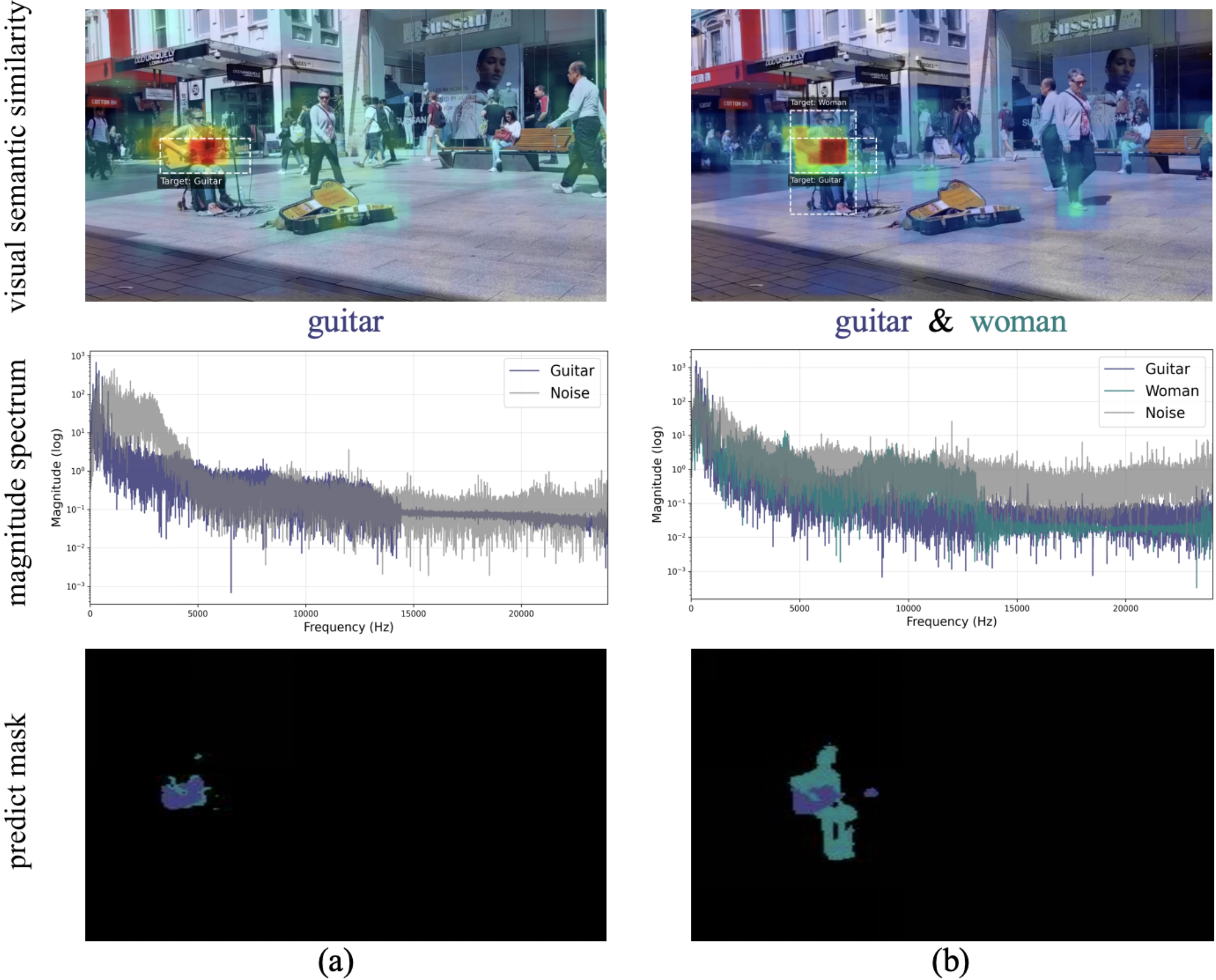}
   \caption{   
   llustration of challenges from intra-modal noise and fine-grained localization. (a) When the target is guitar only, performer–instrument interaction introduces visual interference, while environmental sounds corrupt audio signal, causing blurred boundaries and partial segmentation (\eg, missing guitar head). (b) When the target is to guitar \& woman, visual confusion is alleviated, but audio noise and cross-modal misalignment persist, leading to inaccurate boundaries and incomplete segmentation.}
   \label{fig:1}
\end{figure}

\begin{figure}[h]
   \centering
   \includegraphics[width=0.9\linewidth]{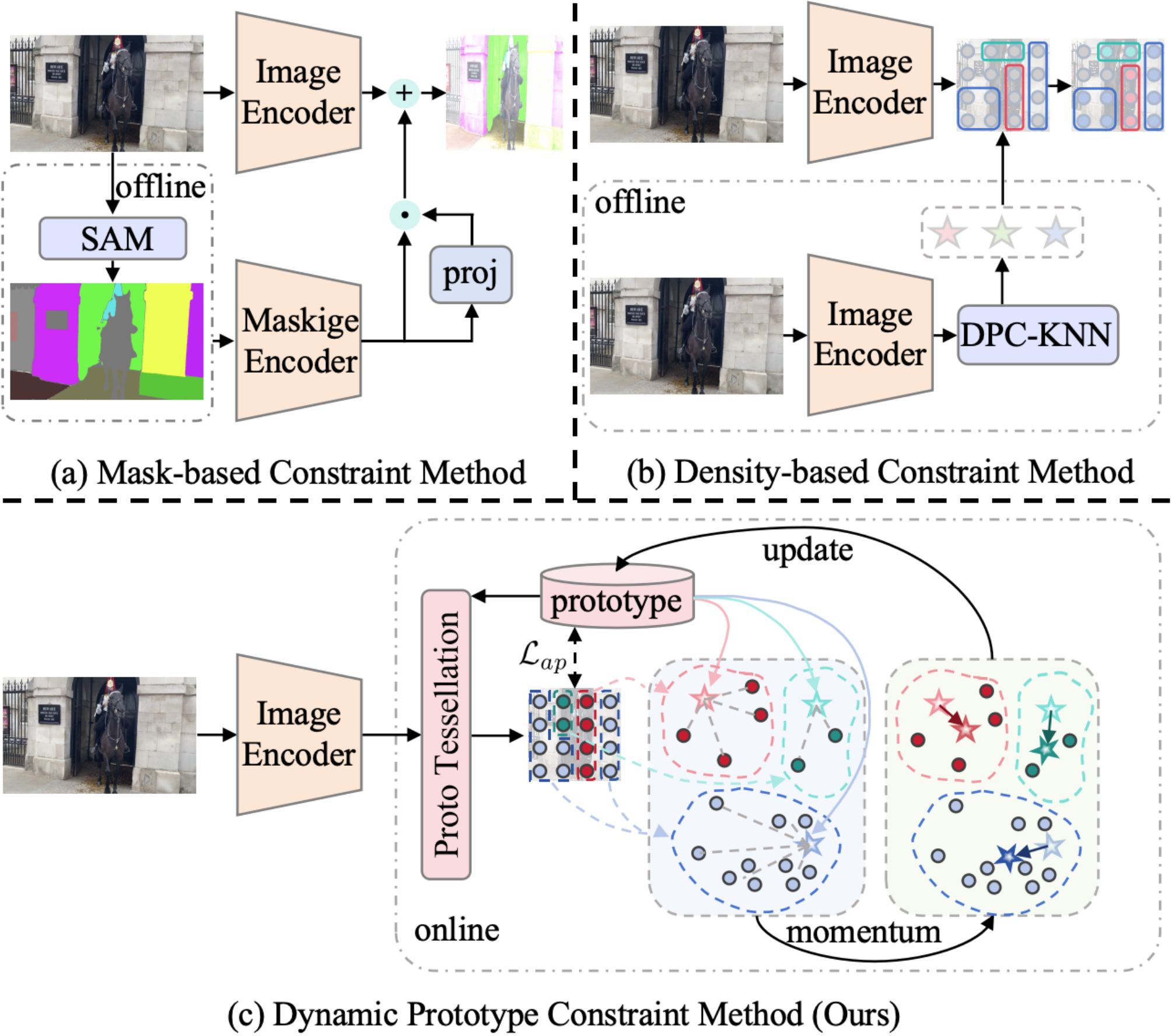}
   \caption{Image denoising methods: (a) The pretrained SAM~\cite{sam} model is employed to generate a prior mask, which is then fused with the original image information; (b) Offline generated prototype centers are introduced to guide pixel clustering within the image; (c) The prototype memory bank is dynamically updated following a momentum strategy based on the input data, and it constrains pixel-wise classification through prototype alignment.}
   \label{fig:proto}
\end{figure}

\section{Introduction}
Audio-Visual Segmentation (AVS)~\cite{AVSBench2022, AVSBench2024} aims to segment frame-wise regions of sounding objects, conditioned on the category of the corresponding audio source. Compared with other audio-visual tasks such as Audio-Visual Event Localization (AVEL)~\cite{ave, psp, sunchao, 10856377} and Sound Source Localization (SSL)~\cite{ssl1, ssl2, ssl3, ssl_ssc}, AVS is more challenging due to its fine-grained pixel-level discrimination. Early studies~\cite{AVSBench2024, AVSegformer, GAVS} primarily focused on cross-modal interactions and spatio-temporal correlations, while overlooking the intra-modal noise within audio and visual streams. As illustrated in \cref{fig:1}, semantic similarities between the target and background in visual scenes, as well as frequency overlaps among multiple audio components, introduce severe intra-modal interference during cross-modal fusion. Furthermore, the lack of fine-grained pixel perception often leads to blurred object boundaries and incomplete or even incorrect segmentation results.

To address the aforementioned challenges, recent studies~\cite{COMBO, VCT, DDESEG, RAVS} introduce prototype learning and prior knowledge (\eg, anchor boxes or masks generated by SAM~\cite{sam}) to suppress intra-modal noise. Others~\cite{CAVP, CPM} adopt contrastive learning strategies to achieve adaptive alignment between audio and visual modalities, thereby mitigating cross-modal interference to some extent. Although these approaches have shown promising results, they often incur additional computational costs and suffer from limited generalization and robustness.

In contrast to prior methods, we present a fully adaptive framework comprising two key components: the Self-Supervised Audio Enhancement (SSAE) and Dynamic Prototype Constraint (DPC). The SSAE module performs intra-modal denoising and preliminary cross-modal alignment, while DPC facilitates fine-grained interaction between modalities.
Inspired by \cite{BYOL, SimSiam, SwAV, DINO}, SSAE treats audio and visual inputs as distinct views and adopts a self-supervised contrastive learning scheme without relying on explicit negative pairs. Unlike conventional uni-modal contrastive learning with shared branch weights, the large semantic gap between modalities hinders direct alignment. To address this, a multi-head cross-modal attention layer is introduced before the projection heads to achieve early semantic alignment, enabling the audio branch to predict visual semantics and suppress irrelevant noise. This design allows SSAE to adaptively enhance correlated features and mitigate intra-modal noise. Moreover, by removing manually constructed pairs or predefined thresholds as in~\cite{CAVP, CPM, RAVS, ICF}, our method improves generalization to in-the-wild data and reduces computational overhead. SSAE thus represents a novel extension of self-supervised contrastive learning to cross-modal representation alignment, distinct from SACL~\cite{ssl_ssc} in SSL tasks that treat single-modal features as dual-branch inputs.

For visual denoising, the DPC module also differs fundamentally from prior-driven approaches~\cite{COMBO, RAVS, DDESEG, VCT} owing to its adaptive design. As illustrated in \cref{fig:proto}, instead of relying on offline priors to guide object localization, DPC dynamically updates a randomly initialized prototype bank during training according to the category distribution within each batch, using momentum and adaptive mechanisms to ensure stable evolution. In the Proto Tessellation module, the prototype bank constrains pixel-wise classification via an Adaptive Prototype loss ($\mathcal{L}_{ap}$). After pixel-level classification, DPC performs audio-visual interaction to obtain class-specific response scores, which are then used to enhance pixels belonging to sounding objects while suppressing irrelevant regions. This pixel-level classification and reinforcement strategy significantly improves the model’s fine-grained discrimination capability and enhances segmentation accuracy.

In summary, our main contributions are as follows:
\begin{itemize}
\item We propose a fully adaptive framework, BYOAVP, to address the prevalent intra-modal interference in AVS tasks.



\item To mitigate cross-modal semantic gaps and noisy audio representations, we propose the SSAE module, which treats dual-modal features as two branches in self-supervised contrastive learning, enabling high-level visual semantics to guide audio denoising.

\item To suppress visual noise and improve fine-grained discrimination, we introduce the DPC module, which maintains a momentum-adaptive prototype bank to constrain pixel-wise classification, and further enhances pixel-level target localization through cross-modal interaction.

\item Extensive experiments demonstrate that our model achieves new state-of-the-art performance across six sub-tasks on two AVS benchmarks.
\end{itemize}

\section{Related Work}
\subsection{Audio-Visual Segmentation}
Audio-Visual Segmentation (AVS) aims to segment sounding objects in a scene by jointly analyzing auditory and visual information. AVSBench~\cite{AVSBench2022} first introduced a framework that performs multimodal feature fusion for pixel-level dense prediction, pioneering the AVS task. Since then, numerous approaches~\cite{AVSegformer, AQFormer, Selm, Catr, AVSC, Bavs} have adopted transformer-based architectures with audio queries to enhance cross-modal interaction. However, these methods generally overlook the intra-modal noise inherently in both audio and visual signals. Recent studies~\cite{COMBO, VCT, DDESEG, RAVS, AuralSAM2} address this issue by incorporating prior knowledge for single-modality denoising. For instance, some works~\cite{COMBO, VCT} utilize prior masks generated by Semantic-SAM~\cite{semantic-sam} to enhance visual category perception, while others~\cite{RAVS, DDESEG} employ offline prototype centers to guide clustering of unimodal features. Meanwhile, contrastive learning strategies have also been introduced to encourage adaptive alignment between audio and visual modalities and alleviate intra-modal noise interference. For example, CAVP~\cite{CAVP} and CPM~\cite{CPM} construct positive–negative sample pairs manually to learn audio-visual correspondence across different video samples, whereas RAVS~\cite{RAVS} assigns positive and negative pairs by setting similarity thresholds derived from audio-visual interactions for single-frame contrastive learning. 
Although these methods show promising results, they incur high computational overhead and limited generalization in complex real-world scenarios.



\subsection{Self-Supervised Contrastive Learning}
Contrastive learning aims to learn discriminative representations by maximizing the consistency between different views of the same instance while distinguishing them from others. However, conventional frameworks~\cite{MoCo, MoCov2, SimCLR, SimCLRv2, cmc, DiffColor} typically rely on explicit negatives or momentum encoders to avoid representation collapse, which increases overall training complexity. To alleviate this dependency, BYOL~\cite{BYOL} and SimSiam~\cite{SimSiam} introduce asymmetric architectures and stop-gradient mechanisms, showing that stable and discriminative features can be learned without explicit negatives. These advances significantly simplify self-supervised frameworks and enhance both training stability and cross-domain generalization. Inspired by these studies, we extend self-supervised contrastive learning to the audio-visual domain and adapt it to multi-modal inputs, effectively addressing the long-standing semantic gap across modalities and intra-modal noise interference.
\begin{figure*}[ht]
   \centering
   \includegraphics[width=0.8\linewidth]{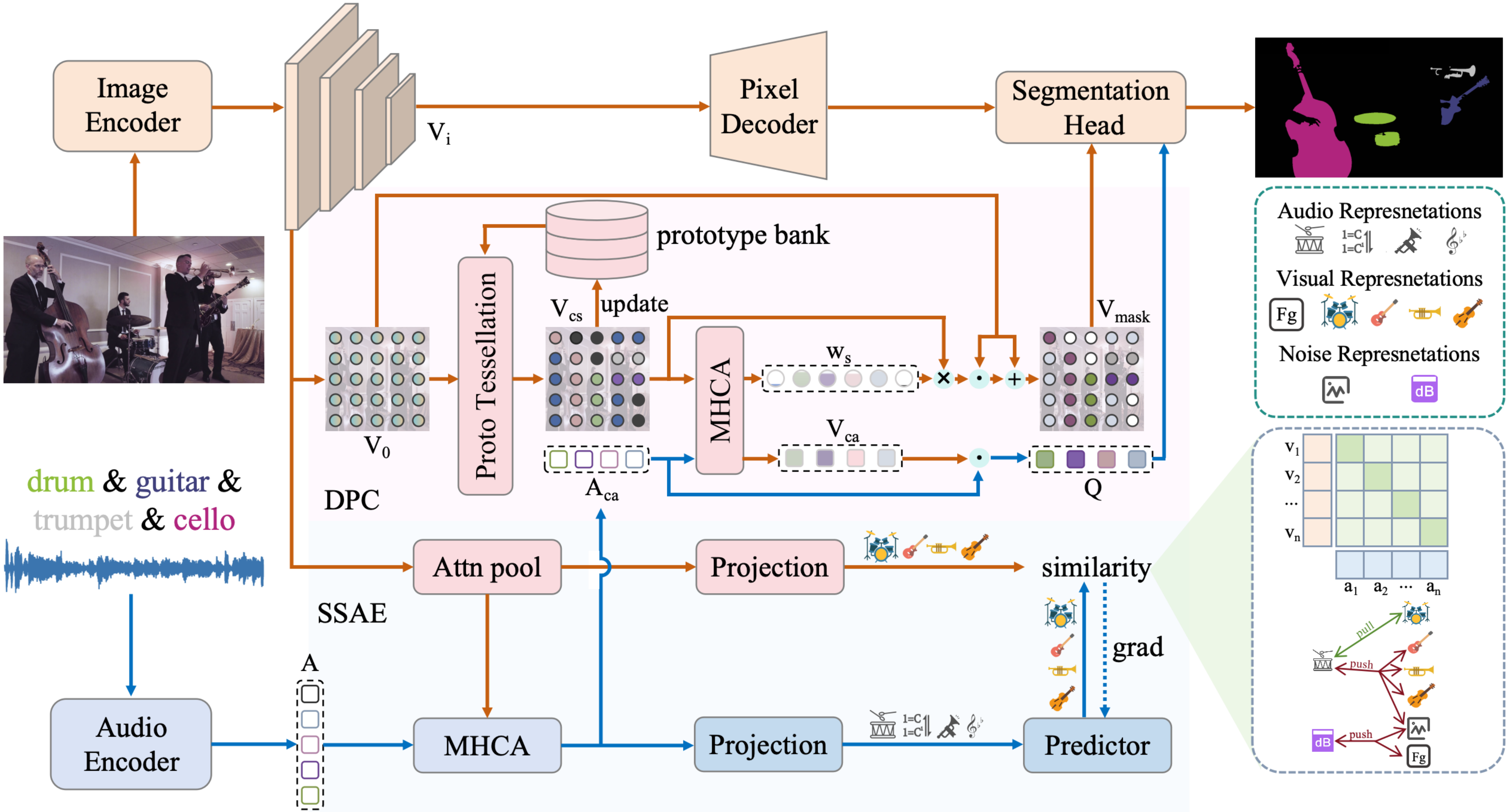}
   \caption{Overall architecture of BYOAVP. After encoding, both image and audio inputs are transformed into feature representations. The first-layer visual features are fed into the DPC module for pixel classification under prototype bank constraints, which are dynamically updated. Meanwhile, the audio features and last-layer visual features undergo cross-modal self-supervised contrastive learning to obtain enhanced audio embeddings. These embeddings interact with the classified visual features to produce target-enhanced mask features and semantically aligned queries, which are then fed into the segmentation head for final prediction.}
   \label{fig:overview}
\end{figure*}
\section{Methodology}
\label{sec:method}
As shown in \cref{fig:overview}, our proposed framework, \textbf{BYOAVP}, addresses AVS challenges with two key modules: Self-Supervised Audio Enhancement (SSAE) and Dynamic Prototype Constraint (DPC). SSAE adopts self-supervised contrastive learning that treats audio and visual features as dual branches, using cross-modal attention to align their semantics by predicting visual outputs from the audio branch. DPC performs pixel-wise classification guided by dynamically updated prototypes and applies cross-modal attention to produce category-specific activations for sounding regions, enhancing discriminative object boundaries.

\subsection{Feature Extraction}
For an input frame, the image encoder encodes it into four-level hierarchical features $\bm{V}_{i}\in\mathbb{R}^{C_{i} \times \frac{H}{2^{i+1}} \times \frac{W}{2^{i+1}} },i \in \{0,1,2,3\}$, in which $C_{i}$, $H$ and $W$ denote the channel dimension of the $i$-th stage output features and the size of image features. For the input audio, we first divide it into 1-second segments to ensure one-to-one correspondence with the video frames. Each segment is then encoded by the audio encoder to obtain the corresponding audio feature, denoted as $\bm{A}\in\mathbb{R}^D$, where $D$ represents feature dimension. 

\subsection{Self-Supervised Audio Enhancement}
\label{method_ssae}
Inspired by \cite{BYOL, SimSiam, SwAV, DINO}, we employ a self-supervised contrastive learning strategy to suppress intra-audio noise while achieving preliminary audio-visual semantic alignment. As shown in \cref{fig:overview}, considering the semantic richness of high-level representations, we employ the last-stage visual feature $\bm{V}_3$ as the input to the visual branch. To capture global contextual semantics, $\bm{V}_3$ is processed by an Attention Pooling Layer. Specifically, the feature map is first flattened into a sequence with shape $(H_3 \times W_3)\times C_3$ and prepended with its mean-pooled representation forming a token sequence of length $(H_3 \times W_3+1)$. A learnable positional embedding of the same length is then added to preserve spatial information, resulting in the embedded feature $\bm{V}_e$. Finally, a multi-head self-attention operation is applied, where the mean token $\bm{\bar{V}}$ (the first element in the sequence) serves as the query, and the entire sequence acts as the key and value. The pooled output $\bm{V}_{pool}$ is obtained as:  
\begin{equation}
  \bm{V}_{pool} = Softmax(\frac{(\bm{\bar{V}}\bm{W}^q_1)(\bm{V}_e\bm{W}^k_1)^T}{\sqrt{C_3}})(\bm{V}_e\bm{W}^v_1),
\end{equation}
where $\bm{W}^q_1$, $\bm{W}^k_1$ and $\bm{W}^v_1$ are learnable linear projections. Subsequently, $\bm{V}_{pool}$ is passed through a linear projection layer to align its dimensionality with that of the audio feature representations, resulting in a shape of $\mathbb{R}^D$.

To reduce the semantic gap between audio and visual modalities, we first feed both modalities into a cross-modal attention module before the features are projected. Considering that the SSAE module ultimately outputs enhanced audio features, we use the audio feature $\bm{A}$ as the key and value of the multi-head cross-modal attention, while $\bm{V}_{pool}$ serves as the query. This design emphasizes the dominance of audio features at this stage, yielding the output $\bm{A}_{ca}$. The formulation is as follows:
\begin{equation}
  \bm{A}_{ca} =\bm{A} + Softmax(\frac{(\bm{V}_{pool}\bm{W}^q_2)(\bm{A}\bm{W}^k_2)^T}{\sqrt{D}})(\bm{A}\bm{W}^v_2),
\end{equation}
where $\bm{W}^q_2$, $\bm{W}^k_2$ and $\bm{W}^v_2$ are learnable linear projections.

After the above preparation, both branches perform standard self-supervised contrastive learning. Despite reduced modality discrepancy, it is unrealistic to assume complete alignment between $\bm{V}_{pool}$ and $\bm{A}_{ca}$. Therefore, unlike conventional contrastive learning, the two branches use independent projection layers without weight sharing.

We adopt the projection layer design from SimSiam~\cite{SimSiam}, which consists of two linear layers, two normalization layers, and a ReLU activation. However, unlike standard computer vision settings, AVS tasks involve video data with substantial memory overhead, making it infeasible to use large batch sizes. Prior studies such as BYOL~\cite{BYOL} suggest that self-supervised contrastive learning is less sensitive to batch size than positive–negative pair–based contrastive methods, yet performance still degrades significantly when the batch size falls below 256. In practice, most AVS models are trained on a single GPU, limiting the batch size to fewer than 8. Under such constraints, relying heavily on Batch Normalization, which is sensitive to batch statistics, inevitably impairs model stability and performance. To mitigate this issue, we replace the final Batch Normalization layer with Group Normalization. The projection layer computation is formulated as follows:
\begin{equation}
  \bm{F} = GN((ReLU(BN(\bm{F}\bm{W^p_1})))\bm{W}^p_2),
\end{equation}
where $\bm{F}$ represents the input feature (\eg, $\bm{V}_{pool}$ or $\bm{A}_{ca}$), $GN(\cdot)$ and $BN(\cdot)$ denote Group and Batch Normalization, and $\bm{W}^p_1$, $\bm{W}^p_2$ are learnable linear projections.

To align the audio embeddings, we allow gradient backpropagation through the audio branch. Thus, the predictor is placed in the audio branch, mapping audio embeddings into the visual semantic space. Specifically, the predictor is an MLP with a structure similar to the projection head but without the final Group Normalization:
\begin{equation}
  \bm{A}_{pred} = (ReLU(BN(\bm{A}_{ca}\bm{W^p_3})))\bm{W}^p_4,
\end{equation}
where $\bm{A}_{pred}$ is the output of the predictor, while $\bm{W}^p_3$ and $\bm{W}^p_4$ are learnable linear projections. 

Subsequently, we implement the $\mathcal{L}_{ssc}$ as a negative cosine similarity loss between $\bm{V}_{pool}$ and $\bm{A}_{pred}$, constraining the entire SSAE process:
\begin{equation}
  \mathcal{L}_{ssc} = -\frac{sg(\bm{V}_{pool}) \odot\bm{A}_{pred}}{||\bm{V}_{pool}||_2||\bm{A}_{pred}||_2},
\end{equation}
where $sg(\cdot)$ and $\odot$ denote the stop-gradient operation and element-wise multiplication. The $\mathcal{L}_{ssc}$ encourages the predicted audio embedding to align with the visual representation. Through gradient backpropagation in the audio branch, it constrains the preceding audio embeddings, suppressing noise factors such as off-screen and background sounds while enhancing the semantics of sounding objects.

\subsection{Dynamic Prototype Constraint}
To achieve pixel-level classification and precise localization, we design an adaptive prototype learning mechanism. As shown in \cref{fig:overview}, the visual input $\bm{V}_0$ from the first stage is flattened into size $N \times C_0$, where $N = H_0 \times W_0$. Then, $\bm{V}_0$ is fed into the Proto Tessellation module for pixel-wise classification. To regularize this process, a randomly initialized prototype bank $\bm{P}_{bank}\in \mathbb{R}^{K\times C_0}$, where $K$ denotes the number of categories, is introduced as an additional input. Specifically, $\bm{P}_{bank}$ interacts with $\bm{V}_0$ through matrix multiplication to produce similarity logits $\bm{P}_{l} \in \mathbb{R}^{N\times K}$, while $\bm{V}_0$ is projected into the class space to obtain $\bm{V}_l \in \mathbb{R}^{N \times K}$. The two are summed and passed through a Gumbel-Softmax~\cite{gumbel_softmax} to achieve differentiable soft pixel assignments. To identify sounding categories, we compute prototype representations weighted by the number of assigned pixels. Concretely, the soft output $\bm{V}_{soft}$ is summed over the spatial dimension $N$ to obtain $\bm{P}_{count}\in \mathbb{R}^{K \times 1}$, denoting the pixel count per prototype. Multiplying $\bm{V}_{soft}$ with $\bm{V}_0$ yields global semantic embeddings of sounding categories, and normalizing by $\bm{P}_{count}$ produces the category-suppressed embedding $\bm{V}_{cs} \in \mathbb{R}^{K \times C_0}$. The process is formulated as follows:
\begin{equation}
  \bm{P}_l = \bm{V}_0 \bm{P}_{bank}^T,
\end{equation}
\begin{equation}
  \bm{V}_{soft} = Softmax((\bm{V}_0\bm{W}^p_5 + \bm{P}_l+\bm{G}_{noise})/\tau),
\end{equation}
\begin{equation}
  \bm{V}_{cs} = \bm{V}_{soft}\bm{V}_0/\bm{P}_{count},
\end{equation}
where $\bm{G}_{noise}$ are \text{i.i.d.} random samples drawn from the $Gumbel(0, 1)$ distribution, $\tau$ denotes the temperature controlling the smoothness of the approximation, and $\bm{W}^p_5$ is a learnable linear projection.

To make the model focus on the sounding categories within the image, we incorporate a multi-head cross-modal attention mechanism into the DPC module. Specifically, the audio embedding $\bm{A}_{ca}$ is projected into the same dimension as $\bm{V}_{cs}$ and used as the query, while $\bm{V}_{cs}$ serves as the key and value. The resulting attention scores $\bm{w}_s$ represent the sounding object categories present in the current video frame. The attended output $\bm{Q} \in \mathbb{R}^{C_0}$ is then multiplied by $\bm{A}_{ca}$ and fed into the segmentation head as its query input. This is mathematically formulated as:
\begin{equation}
  \bm{w}_s = Softmax((\bm{A}_{ca}\bm{W}^q_3)(\bm{V}_{cs}\bm{W}^k_3)^T),
\end{equation}
\begin{equation}
  \bm{Q} = \bm{A}_{ca} \odot (\bm{w}_s (\bm{V}_{cs}\bm{W}^v_3)),
\end{equation}
where $\bm{W}^q_3$, $\bm{W}^k_3$ and $\bm{W}^v_3$ are learnable linear projections.


To further enhance the corresponding sounding regions, the attention scores are multiplied by $\bm{V}_{soft}$ and reshaped to produce pixel-wise category weight maps $\bm{M} \in \mathbb{R}^{N \times 1}$. These weights are reshaped and multiplied with $\bm{V}_0$, and the result is added back to $\bm{V}_0$ to obtain the enhanced mask feature $\bm{V}_{mask}$. Finally, $\bm{V}_{mask}$ is reshaped into $C_0 \times H_0 \times W_0$ and fed into the segmentation head to guide mask prediction. This process is formulated as:
\begin{equation}
  \bm{M} = Reshape(\bm{w}_s\bm{V}_{soft}),
\end{equation}
\begin{equation}
  \bm{V}_{mask} = Reshape(\bm{V}_0 + \bm{V}_0 \odot\bm{M}).
\end{equation}

At the end of the DPC computation, the randomly initialized prototype bank is also dynamically updated during training. As illustrated in \cref{fig:proto} (c), the prototype bank updates its semantic centers based on the spatial semantic positions of pixels in $\bm{V}_{cs}$ using a momentum-based strategy. Specifically, to maintain stability while adaptively adjusting the update strength according to the proportion of each sounding category in the current frame, we introduce two weighting factors: $\bm{w}_m$ and $\bm{w}_{ada}$. Here, $\bm{w}_m$ denotes the momentum coefficient controlling the temporal smoothness of prototype updates, while $\bm{w}_{ada}$ represents the adaptive category weight that scales the update strength based on the category-wise pixel distribution. Given the pixel count matrix $\bm{P}_{count} \in \mathbb{R}^K$, $\bm{w}_{ada}$ is computed as follows:
\begin{equation}
  \bm{w}_{ada} = (1.0-\bm{w}_m)(1.0+\bm{P}_{count}/N),
\end{equation}
where $N$ denotes the total number of pixels.

To stabilize Gumbel-Softmax assignments and improve the interpretability and discriminability of prototypes, we introduce an Adaptive Prototype loss ($\mathcal{L}_{ap}$) to regularize this process. Specifically, the similarity logits $\bm{P}_l$ are amplified and passed through a Log-Softmax function to obtain the predicted prototype distribution $\bm{P}_{log}$. The pseudo label $\bm{V}_{soft}$ weights $\bm{P}_{log}$ via element-wise multiplication, and the result is summed over the prototype dimension $K$. The mean of the negative sum is then taken as the final loss:
\begin{equation}
  \bm{P}_{log}^{(n,k)} = \lambda_{s} \bm{P}_l^{(n,k)} - \log(\sum^K_{j=1}\exp{(\lambda_{s} \bm{P}_l^{(n,j)})}) ,
\end{equation}
\begin{equation}
  \mathcal{L}_{ap} = -\frac{1}{N}\sum^N_{n=1}\sum^K_{k=1}\bm{V}_{soft}^{(n,k)}\odot\bm{P}_{log}^{(n,k)},
\end{equation}
where $n$ indexes pixels, $k$ indexes prototypes, and $\lambda_{s} = 10$ is a hyper-parameter used to amplify $\bm{P}_l$.

\subsection{Loss Function}
The overall training objective comprises three parts: segmentation loss, self-supervised contrastive loss, and adaptive prototype loss. The segmentation loss combines BCE, Dice~\cite{dice}, and Focal~\cite{focal} losses. The total loss is formulated as follows:
\begin{equation}
  \mathcal{L}_{seg} = \mathcal{L}_{bce}+ \mathcal{L}_{dice}+  \lambda_{f} \mathcal{L}_{focal},
\end{equation}
\begin{equation}
  \mathcal{L} = \lambda_{seg}\mathcal{L}_{seg}+ \lambda_{ssc}\mathcal{L}_{ssc}+ \lambda_{ap} \mathcal{L}_{ap},
\end{equation}
where $\lambda_f$, $\lambda_{seg}$, $\lambda_{ssc}$ and $\lambda_{ap}$ are all hyper-parameters used to balance the loss function.
\newcolumntype{P}[1]{>{\centering\arraybackslash}p{#1}} 
\newcolumntype{L}[1]{>{\raggedright\arraybackslash}p{#1}}

\begin{table*}[t]
\centering
\fontsize{8pt}{9pt}\selectfont
\caption{Quantitative comparisons on AVSBench dataset~\cite{AVSBench2022, AVSBench2024} with Transformer backbone. Best results in \textbf{Bold}, while second best \underline{underlined}. Methods marked with * leverage prior knowledge, such as SAM-generated masks or offline prototype centers.}
\label{tab:avs}
\begin{tabular}{L{3.2cm}|P{1.3cm}|P{1.3cm}|P{0.95cm}P{0.6cm}P{0.6cm}|P{0.95cm}P{0.6cm}P{0.6cm}|P{0.95cm}P{0.6cm}P{0.6cm}}
\toprule
\multirow{2}{*}{\textbf{Method}} & 
\multirow{2}{*}{\makecell{\textbf{Visual}\\\textbf{Backbone}}} & 
\multirow{2}{*}{\makecell{\textbf{Audio}\\\textbf{Backbone}}} &
\multicolumn{3}{c|}{\textbf{AVS-Object-S4}} &
\multicolumn{3}{c|}{\textbf{AVS-Object-MS3}} &
\multicolumn{3}{c}{\textbf{AVS-Semantic}} \\
& & & $\mathcal{J} \& \mathcal{F}$ $\uparrow$ & $\mathcal{J}$ $\uparrow$ & $\mathcal{F}$ $\uparrow$ &
$\mathcal{J} \& \mathcal{F}$ $\uparrow$ &
$\mathcal{J}$ $\uparrow$ & $\mathcal{F}$ $\uparrow$ &
$\mathcal{J} \& \mathcal{F}$ $\uparrow$ &
$\mathcal{J}$ $\uparrow$ & $\mathcal{F}$ $\uparrow$ \\
\midrule
AVSBench~\cite{AVSBench2022} \graytext{[ECCV'22]} & PVT-v2 & VGGish & 83.3 & 78.7 & 87.9 & 59.3 & 54.0 & 64.5 & 32.5 & 29.8 & 35.2 \\
ECMVAE~\cite{ECMVAE} \graytext{[ICCV'23]} & PVT-v2 & VGGish & 85.9 & 81.7 & 90.1 & 64.3 & 57.8 & 70.8 & -- & -- & -- \\
AVSC~\cite{AVSC} \graytext{[ACM MM'23]} & Swin-B & VGGish & 85.0 & 81.3 & 88.6 & 62.6 & 59.5 & 65.8 & -- & -- & -- \\
CATR~\cite{Catr} \graytext{[ACM MM'23]} & PVT-v2 & VGGish & 87.9 & 84.4 & 91.3 & 68.6 & 62.7 & 74.5 & 35.7 & 32.8 & 38.5 \\
AQFormer~\cite{AQFormer} \graytext{[IJCAI'23]} & PVT-v2 & VGGish & 85.5 & 81.6 & 89.4 & 67.5 & 62.2 & 72.7 & -- & -- & -- \\
AVSegFormer~\cite{AVSegformer} \graytext{[AAAI'24]} & PVT-v2 & VGGish & 86.0 & 82.1 & 89.9 & 63.9 & 58.4 & 69.3 & 39.4 & 36.7 & 42.0\\
MUTR~\cite{MUTR} \graytext{[AAAI'24]} & Swin-B & VGGish & 85.7 & 81.6 & 89.7 & 68.8 & 64.0 & 73.5 & -- & -- & -- \\
AVSBG~\cite{AVSBG} \graytext{[AAAI'24]} & PVT-v2 & VGGish & 86.1 & 81.7 & 90.4 & 61.0 & 55.1 & 66.8 & -- & -- & -- \\
GAVS~\cite{GAVS} \graytext{[AAAI'24]} & ViT-B & VGGish & 85.2 & 80.1 & 90.2 & 70.6 & 63.7 & 77.4 & -- & -- & -- \\
BAVS*~\cite{Bavs} \graytext{[TMM'24]} & Swin-B & Beats & 86.2 & 82.7 & 89.8 & 62.8 & 59.6 & 65.9 & 35.6 & 33.6 & 37.5 \\
QDFormer~\cite{QDformer} \graytext{[CVPR'24]} & Swin-T & VGGish & 83.9 & 79.5 & 88.2 & 64.1 & 62.0 & 66.1 & -- & 53.4 & -- \\
COMBO*~\cite{COMBO} \graytext{[CVPR'24]} & PVT-v2 & VGGish & 88.3 & 84.7 & 91.9 & 65.2 & 59.2 & 71.2 & 44.1 & 42.1 & 46.1 \\
UFE~\cite{UFE} \graytext{[CVPR'24]} & PVT-v2 & VGGish & 86.8 & 83.2 & 90.4 & 66.5 & 62.0 & 70.9 & -- & -- & -- \\
AAVS~\cite{AAVS} \graytext{[ECCV'24]}& Swin-B & VGGish & 87.3 & 83.2 & 91.3 & 72.5 & 67.3 & 77.6 & 50.9 & 48.5 & 53.2 \\
BiasAVS~\cite{AVSBias} \graytext{[ACM MM'24]}& Swin-B & VGGish & 88.2 & 83.3 & 93.0 & 74.0 & 67.2 & 80.8 & 47.2 & 44.4 & 50.0 \\
SelM~\cite{Selm} \graytext{[ACM MM'24]}& PVT-v2 & VGGish & 87.4 & 83.5 & 91.2 & 65.8 & 60.3 & 71.3 & 44.1 & 41.3 & 46.9 \\
DiffusionAVS~\cite{DiffusionAVS} \graytext{[TIP'25]} & PVT-v2 & VGGish & 85.7 & 81.4 & 90.0 & 64.6 & 58.2 & 70.9 & 40.6 & 38.1 & 43.0\\
AVS-Mamba~\cite{Avs-mamba} \graytext{[TMM'25]} & PVT-v2 & VGGish & 88.8 & 85.0 & 92.6 & 73.7 & 68.6 & 78.8 & 42.4 & 39.7 & 45.1 \\
VCT*~\cite{VCT} \graytext{[CVPR'25]} & Swin-B & VGGish & 88.5 & 84.7 & 92.3 & 73.4 & 67.5 & 79.3 & 50.4 & 47.9 & 52.9 \\
RAVS*~\cite{RAVS} \graytext{[CVPR'25]} & MIT-B5 & HTSAT & 93.5 & \underline{93.1} & 93.8 & 76.4 & 70.6 & 82.1 & 65.7 & 60.8 & 70.6 \\
DDESEG*~\cite{DDESEG} \graytext{[CVPR'25]} & MIT-B5 & HTSAT & \underline{94.2} & 92.4 & \underline{95.9} & 77.9 & 72.3 & 83.4 & \underline{67.9} & \underline{63.4} & \underline{72.3} \\
ICF*~\cite{ICF} \graytext{[ICCV'25]} & PVT-v2 & VGGish & 90.1 & 86.6 & 93.5 & 69.9 & 64.4 & 75.4 & 48.2 & 45.0 & 51.3 \\
AuralSAM2*~\cite{AuralSAM2} \graytext{[arXiv'25]} & Hiera-l & VGGish & 90.0 & 86.6 & 93.3 & \underline{79.9} & \underline{75.6} & \textbf{84.1} & 53.3 & 50.6 & 56.0 \\
AV2T-SAM*~\cite{AV2T-SAM} \graytext{[ICIP'25]} & CLIP & CLAP & 89.6 & 86.7 & 92.4 & 73.7 & 69.7 & 77.7 & -- & -- & -- \\
\rowcolor{gray!30}
\textbf{BYOAVP (ours)} & Swin-B & CLAP & \textbf{95.0} & \textbf{93.2} & \textbf{96.7} & \textbf{80.5} & \textbf{77.4} & \underline{83.6} & \textbf{69.6} & \textbf{64.7} & \textbf{74.5} \\
\bottomrule
\end{tabular}
\end{table*}
\section{Experiments}

\subsection{Experimental Setup}

\par
\noindent\textbf{Datasets.}
We conduct experiments on two datasets: \
\begin{itemize}
\item \textbf{AVSBench}~\cite{AVSBench2022, AVSBench2024} consists of three sub-tasks. The single sound source (S4) and multi sound sources (MS3) sub-tasks use 5-frame clips as input, covering 23 sounding categories with 4,932 and 424 videos, respectively. Both require binary segmentation of sounding objects. In contrast, the AVS-Semantic (AVSS) sub-task extends the dataset to 10-frame inputs with 12,356 videos from 71 sound categories, where the goal is semantic-level segmentation of sounding objects.

\item \textbf{VPO}~\cite{CAVP} is a synthetic benchmark for audio-visual segmentation, constructed by pairing single-frame images and their semantic masks from COCO~\cite{coco} with three-second audio clips from VGGSound~\cite{vggsound}. It covers 21 sound-producing categories and consists of three sub-tasks: single-source (VPO-SS, 12,202 samples), multi-source (VPO-MS, 9,817 samples), and multi-source multi-instance (VPO-MSMI, 12,855 samples).
\end{itemize}

\par
\noindent\textbf{Metrics.}
We evaluate performance using two standard metrics: the Jaccard index ($\mathcal{J}$) and the F-score ($\mathcal{F}$).

\par
\noindent\textbf{Implementation Details.}
We adopt the Swin Transformer (Swin-B)~\cite{swin-trans}, pretrained on ImageNet-21K~\cite{ImageNet-21k}, as our visual encoder. For the audio branch, we employ the HTSAT~\cite{HTSAT} from CLAP~\cite{CLAP2023} as our audio feature extractor and directly utilize the official pretrained weights, which were trained on 22 audio-related datasets, such as AudioSet~\cite{AudioSet}. Following \cite{COMBO, ICF}, we employ a Multi-Scale Deformable Attention Transformer (Deformable DETR)~\cite{Deformable-DETR} as the pixel decoder. The loss coefficients $\lambda_{seg}$, $\lambda_{ssc}$, and $\lambda_{ap}$ are empirically set to 5, 0.1, and 1, respectively (see the \textit{supplementary material} for details). We set the batch size to 4 during training and employ AdamW as the optimizer, with an initial learning rate of 1e-4 and weight decay of $0.05$. 

\begin{figure*}[ht]
   \centering
   \includegraphics[width=1.0\linewidth]{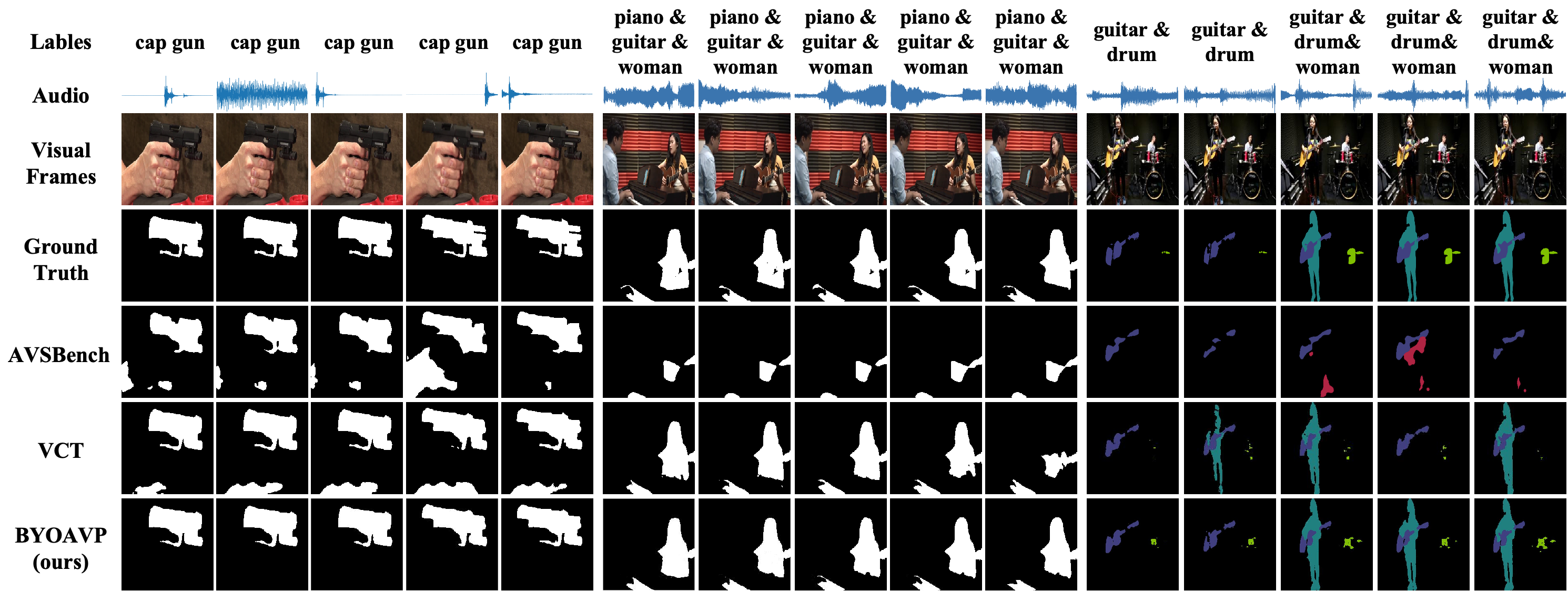}

   \caption{Qualitative comparison of our BYOAVP with AVSBench~\cite{AVSBench2022, AVSBench2024} and VCT~\cite{VCT} on the AVSBench dataset. Representative examples are drawn from the AVS-Object-S4 (left), AVS-Object-MS3 (middle), and AVS-Semantic (right) test sets.}
   \label{fig:ablation}
\end{figure*}
\begin{table*}[t]
\centering
\fontsize{8pt}{9pt}\selectfont
\caption{Quantitative comparisons on VPO dataset~\cite{CAVP}. Best results in \textbf{Bold}, while second best \underline{underlined}. The * indicates that the model utilizes additional prior information, such as offline-generated prototype centers.}
\label{tab:vpo}
\begin{tabular}{L{3.2cm}|P{1.15cm}|P{1.15cm}|P{1.2cm}|P{0.7cm}P{0.5cm}P{0.5cm}|P{0.7cm}P{0.5cm}P{0.5cm}|P{0.7cm}P{0.5cm}P{0.5cm}}
\toprule
\multirow{2}{*}{\textbf{Method}} & 
\multirow{2}{*}{\makecell{\textbf{Visual}\\\textbf{Backbone}}} & 
\multirow{2}{*}{\makecell{\textbf{Audio}\\\textbf{Backbone}}} &
\multirow{2}{*}{\makecell{\textbf{Trainable}\\\textbf{Params}}} &
\multicolumn{3}{c|}{\textbf{VPO-SS}} &
\multicolumn{3}{c|}{\textbf{VPO-MS}} &
\multicolumn{3}{c}{\textbf{VPO-MSMI}} \\
& & & & $\mathcal{J} \& \mathcal{F}$$\uparrow$ & $\mathcal{J}$$\uparrow$ & $\mathcal{F}$$\uparrow$ &
$\mathcal{J} \& \mathcal{F}$$\uparrow$ &
$\mathcal{J}$$\uparrow$ & $\mathcal{F}$$\uparrow$ &
$\mathcal{J} \& \mathcal{F}$$\uparrow$ &
$\mathcal{J}$$\uparrow$ & $\mathcal{F}$$\uparrow$ \\
\midrule
AVSBench~\cite{AVSBench2022} \graytext{[ECCV'22]} & PVT-v2 & VGGish & 101.32M & 44.6 & 41.6 & 47.6 & 45.7 & 42.3 & 49.1 & 43.2 & 40.0 & 46.3 \\
AVSegFormer~\cite{AVSegformer} \graytext{[AAAI'24]} & PVT-v2 & VGGish & 186.05M & 46.0 & 43.8 & 48.1 & 43.7 & 47.3 & 40.1 & 50.0 & 47.2 & 52.7 \\
CAVP~\cite{CAVP} \graytext{[CVPR'24]} & ResNet50 & VGGish & 119.79M & 67.0 & 58.8 & 75.2 & 61.3 & 53.2 & 69.4 & 56.5 & 48.2 & 64.8 \\
CPM~\cite{CPM} \graytext{[ECCV'24]} & ResNet50 & VGGish & -- & 73.5 & 67.1 & 79.9 & 72.9 & 65.9 & 79.9 & 68.1 & 60.6 & 75.6 \\
DDESEG*~\cite{DDESEG} \graytext{[CVPR'25]} & MIT-B5 & HTSAT & 103.84M & 74.4 & 67.6 & 81.2 & 74.3 & 67.6 & 81.0 & 68.4 & 62.1 & 74.7 \\
RAVS*~\cite{RAVS} \graytext{[CVPR'25]} & MIT-B5 & HTSAT & 103.84M & \underline{75.0} & \underline{68.0} & \underline{81.9} & 73.5 & 67.0 & 80.0 & 69.3 & 61.9 & \underline{76.7} \\
\rowcolor{gray!15}
\textbf{BYOAVP (ours)} & MIT-B5 & HTSAT & 134.15M & 74.5 & 67.6 & 81.4 & \underline{75.5} & \underline{69.8} & \underline{81.2} & \underline{70.6} & \underline{64.4} & \underline{76.7} \\
\rowcolor{gray!30}
\textbf{BYOAVP (ours)} & Swin-B & HTSAT & 137.15M & \textbf{75.5} & \textbf{68.8} & \textbf{82.1} & \textbf{77.2} & \textbf{71.9} & \textbf{82.4} & \textbf{71.5} & \textbf{65.8} & \textbf{77.2} \\
\bottomrule
\end{tabular}
\end{table*}
\subsection{Quantitative Comparison}
\par
\noindent\textbf{Performance on AVSBench Dataset.}
As shown in \cref{tab:avs}, with the transformer-based visual encoder, we conduct a comprehensive comparison between our model and previous state-of-the-art methods on the AVSBench dataset. Our approach consistently achieves the best overall performance across all three sub-tasks. Specifically, on the AVS-Object-S4 task, BYOAVP attains $\mathcal{J} \& \mathcal{F}$ scores of 95.0 \textit{vs.} 94.2 (second-best, DDESEG~\cite{DDESEG}), and on the AVS-Object-MS3 task, 80.5 \textit{vs.} 79.9 (second-best, AuralSAM2~\cite{AuralSAM2}). For the AVS-Semantic task, our model surpasses DDESEG by 1.7 in $\mathcal{J} \& \mathcal{F}$. Notably, DDESEG relies on offline-generated prototype centers and bounding boxes, while AuralSAM2 leverages SAM2~\cite{sam2} with a substantially higher input resolution ($1024 \times 1024$) compared to $224 \times 224$ used by other models. Despite these advantages, our framework still outperforms all competitors across the board. Due to space limitations, the results based on the ResNet-50~\cite{resnet50} backbone are provided in the \textit{supplementary material}.

\par
\noindent\textbf{Performance on VPO Dataset.}
As shown in \cref{tab:vpo}, we further compare our model with state-of-the-art methods on the VPO dataset. Our approach achieves the best performance across all three VPO sub-tasks, surpassing the second-best methods in $\mathcal{J} \& \mathcal{F}$ by 0.5, 2.9, and 2.2 on the SS, MS, and MSMI tasks, respectively. The notable gains on the MS and MSMI tasks demonstrate the superior capability of our model in multi-object perception and foreground–background disambiguation within the same category. Although our model contains slightly more trainable parameters than some transformer-based counterparts, methods such as RAVS~\cite{RAVS} and DDESEG~\cite{DDESEG} depend on additional prior information, while our training process is entirely online. However, the additional 33M is acceptable considering the 88M parameters of the visual encoder, and our total parameter count remains significantly lower than that of AVSegFormer~\cite{AVSegformer}.
\subsection{Qualitative Comparison}
We conduct extensive visual comparisons with VCT~\cite{VCT} and AVSBench~\cite{AVSBench2024} across all three AVSBench tasks, as shown in \cref{fig:ablation}. Our model consistently excels at accurately localizing sounding objects and generating masks with sharper boundaries. For instance, in the AVS-Object-S4 example, our method correctly identifies the sounding cap gun, whereas the other methods mistakenly respond to the cushion beneath the shooter’s hand. In the challenging AVS-Semantic scenario, both VCT and AVSBench fail to detect the ``woman'' and nearly overlook the ``drum'', while our model successfully captures all sound sources, demonstrating superior multi-object perception and precise localization. More visualizations and examples can be found in the \textit{supplementary material}.
\subsection{Ablation Study}
\label{exp:ablation}
We conduct comprehensive ablation studies to validate the effectiveness of each component in our model. Focusing on the most challenging multi-sound-source cases, evaluations are performed on AVS-Object-MS3 and AVS-Semantic.

\par
\begin{figure}[t]
   \centering
   \includegraphics[width=0.85\linewidth]{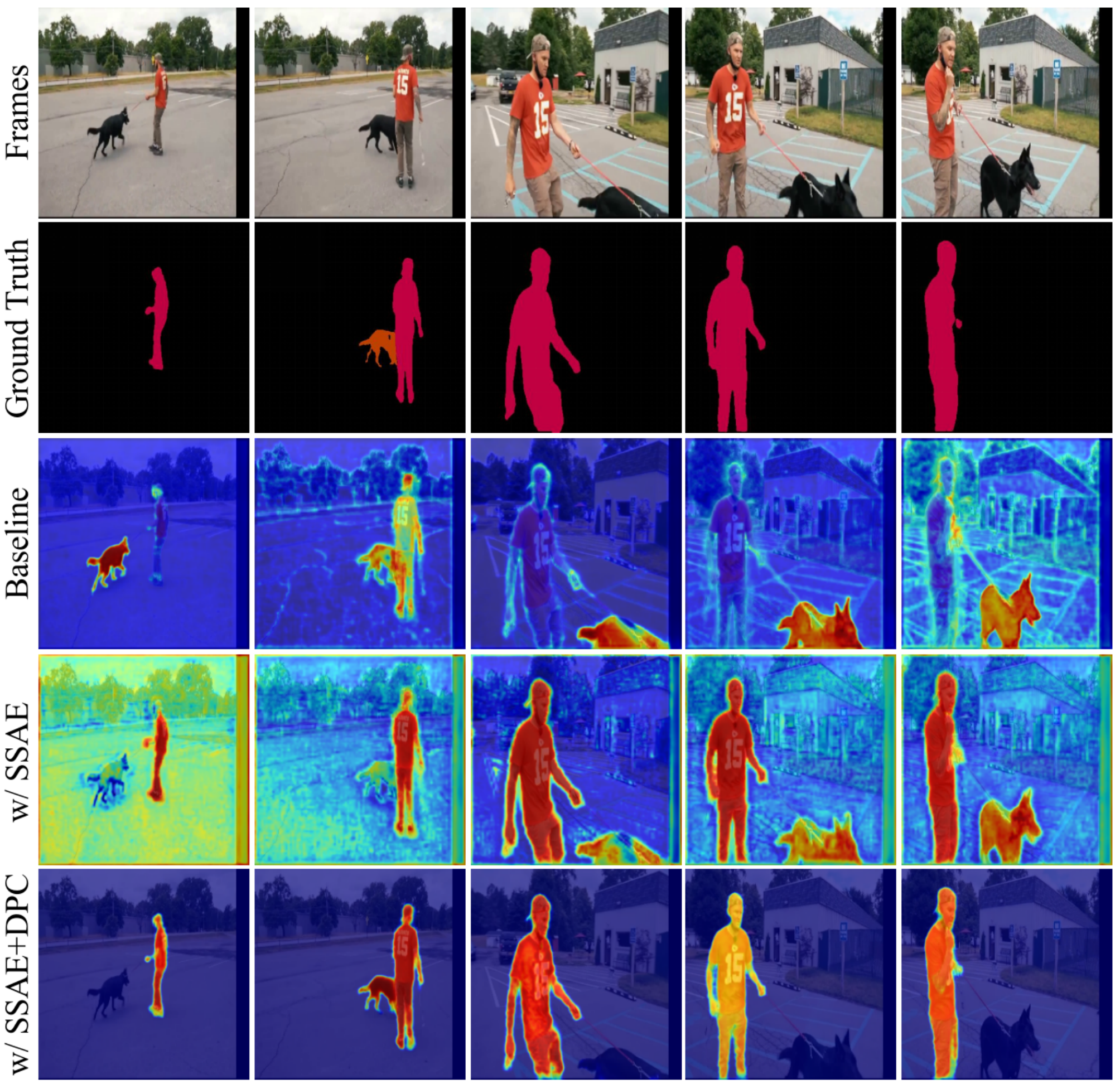}

   \caption{Qualitative comparison of our proposed components. The predicted mask heatmaps show that, from top to bottom, the model progressively concentrates on sounding objects while effectively suppressing surrounding noise.}
   \label{fig:heatmap}
\end{figure}

\noindent\textbf{Component-wise Analysis.}
Our baseline includes an audio encoder, image encoder, pixel decoder, and segmentation head. We progressively add the proposed modules to evaluate their effects. As shown in \cref{tab:ab_all}, introducing either SSAE or DPC yields clear performance gains. Integrating both raises the $\mathcal{J}\&\mathcal{F}$ scores by 8.02 and 7.72 on AVS-Object-MS3 and AVS-Semantic, showing the benefit of joint audiovisual denoising and fine-grained audio–pixel interaction. The heatmaps in \cref{fig:heatmap} further demonstrate improved noise suppression and object localization.

\begin{table}[t]
\centering
\fontsize{8pt}{9pt}\selectfont
\caption{Component effectiveness study of BYOAVP.}
\begin{tabular}{L{1.4cm}|P{0.7cm}P{0.5cm}P{0.6cm}|P{0.7cm}P{0.5cm}P{0.6cm}}
\toprule
\multirow{2}{*}{\textbf{Module}} & 
\multicolumn{3}{c|}{\textbf{AVS-Object-MS3}} &
\multicolumn{3}{c}{\textbf{AVS-Semantic}} \\
& $\mathcal{J} \& \mathcal{F}$$\uparrow$ &
$\mathcal{J}$$\uparrow$ & $\mathcal{F}$$\uparrow$ &
$\mathcal{J} \& \mathcal{F}$$\uparrow$ &
$\mathcal{J}$$\uparrow$ & $\mathcal{F}$$\uparrow$ \\
\midrule
Baseline & 72.48 & 67.58 & 77.38 & 61.89 & 58.24 & 65.53 \\
SSAE & 76.22 & 72.93 & 79.50 & 65.08 & 61.21 & 68.95 \\
DPC & 77.98 & 74.88 & 81.08 & 67.40 & 63.01 & 71.79 \\
\rowcolor{gray!30}
SSAE+DPC & \textbf{80.50} & \textbf{77.43} & \textbf{83.57} & \textbf{69.61} & \textbf{64.68} & \textbf{74.54} \\
\bottomrule
\end{tabular}
\label{tab:ab_all}
\end{table}

\par
\begin{figure}[t]
   \centering
   \includegraphics[width=0.95\linewidth]{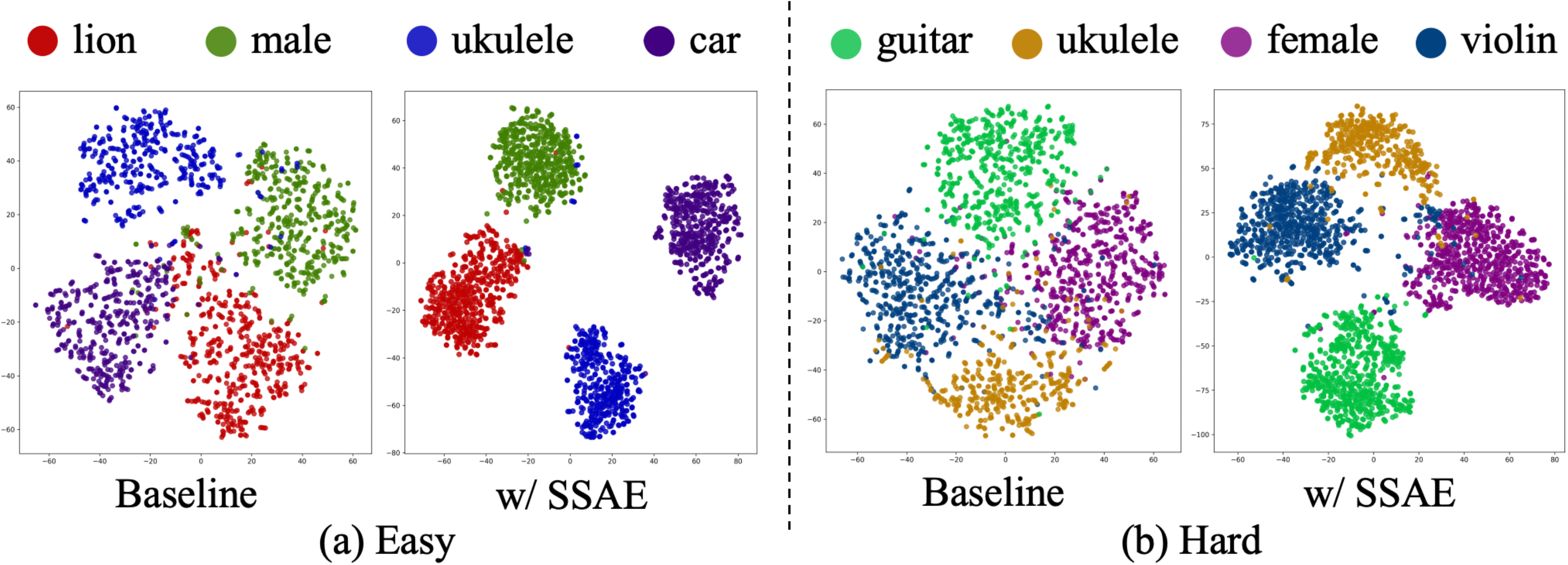}

   \caption{t-SNE visualization of audio features. Easy indicates that the selected categories do not share the same superclass (\eg, animals, instruments), while hard denotes categories belonging to the same superclass or producing acoustically similar sounds.}
   \label{fig:tsne}
\end{figure}

\noindent\textbf{Analysis of Self-Supervised Audio Enhancement.}
We further conduct an in-depth analysis of the SSAE module to better understand how self-supervised contrastive learning can be effectively applied in the presence of cross-modal inputs. To address the challenges posed by the cross-modal semantic gap and the instability induced by small batch sizes, we investigate two key design choices: (1) whether cross-modal interaction should be performed before the single-branch embedding projection, and (2) which normalization strategy is most suitable at the end of the projection layer. As shown in \cref{tab:ab_ssae}, enabling SSAE with only CA, BN, or GN performs worse than disabling SSAE, since CA alone introduces modality-specific bias without stabilizing the embedding space, while normalization alone cannot bridge the semantic gap. These observations confirm that effective self-supervised learning requires the joint use of pre-mapping cross-modal interaction and post-mapping normalization. Furthermore, the comparison between BN and GN reveals that GN yields more stable and consistent learning dynamics, improving $\mathcal{J}\&\mathcal{F}$ by 1.65 and 0.85 on the AVS-Object-MS3 and AVS-Semantic tasks, respectively. This advantage is primarily due to GN’s independence from batch statistics, which effectively mitigates the optimization instability caused by small batch training. As visualized in \cref{fig:tsne}, combining cross-modal interaction with GN enables the SSAE module to generate more compact and separable audio representations, enhancing multi-category perception and fine-grained class discrimination. 

\begin{table}[t]
\centering
\fontsize{8pt}{9pt}\selectfont
\caption{Ablation study of the SSAE module on the AVSBench dataset. “$\checkmark$” and “$\times$” denote whether a component is enabled. CA represents the Cross-Attention layer, while BN and GN denote BatchNorm and GroupNorm applied after the projection.}
\begin{tabular}{P{0.25cm}P{0.25cm}P{0.35cm}|P{0.7cm}P{0.5cm}P{0.6cm}|P{0.7cm}P{0.5cm}P{0.6cm}}
\toprule
\multicolumn{3}{c|}{\textbf{Strategy}} & 
\multicolumn{3}{c|}{\textbf{AVS-Object-MS3}} &
\multicolumn{3}{c}{\textbf{AVS-Semantic}} \\
\midrule
CA & BN & GN &
$\mathcal{J} \& \mathcal{F}$$\uparrow$ &
$\mathcal{J}$$\uparrow$ & $\mathcal{F}$$\uparrow$ &
$\mathcal{J} \& \mathcal{F}$$\uparrow$ &
$\mathcal{J}$$\uparrow$ & $\mathcal{F}$$\uparrow$ \\
\midrule
$\times$ & $\times$ & $\times$ & 77.98 & 74.88 & 81.08 & 67.40 & 63.01 & 71.79 \\
$\checkmark$ & $\times$ & $\times$ & 77.54 & 74.23 & 80.85 & 66.60 & 62.17 & 71.02 \\
$\times$ & $\checkmark$ & $\times$ & 76.95 & 73.59 & 80.31 & 66.09 & 61.33 & 70.84 \\

$\times$ & $\times$ & $\checkmark$ & 77.77 & 74.75 & 80.79 & 67.38 & 62.79 & 71.96 \\
$\checkmark$ & $\checkmark$ & $\times$ & 78.85 & 75.41 & 82.29 & 68.76 & 63.72 & 73.80 \\
\rowcolor{gray!30}
$\checkmark$ & $\times$ & $\checkmark$ & \textbf{80.50} & \textbf{77.43} & \textbf{83.57} & \textbf{69.61} & \textbf{64.68} & \textbf{74.54} \\
\bottomrule
\end{tabular}
\label{tab:ab_ssae}
\end{table}

\section{Conclusion}
In this work, we propose BYOAVP, a novel audio-visual segmentation framework that effectively mitigates intra-modal noise and cross-modal semantic gaps without relying on prior information or manually constructed sample pairs. The framework consists of two core components: the Self-Supervised Audio Enhancement (SSAE) module, which innovatively introduces self-supervised contrastive learning into cross-modal tasks for audio feature denoising and enhancement; and the Dynamic Prototype Constraint (DPC) module, which maintains class prototypes via momentum updates to regularize pixel-level classification and reinforce semantically consistent regions. Extensive experiments on two AVS benchmarks and six sub-tasks demonstrate that BYOAVP achieves new state-of-the-art performance, validating its robustness and generalization capability.

\section*{Acknowledgements}

This work is partially supported by the National Key Research and Development Program of China (2023YFC2705702) and the LIESMARS Special Research Funding. This work was also supported by WHU-Kingsoft Joint Lab. The numerical calculations in this paper have been done on the supercomputing system in the Supercomputing Center of Wuhan University.

{
    \small
    \bibliographystyle{ieeenat_fullname}
    \bibliography{main}
}


\end{document}